\documentclass[letterpaper, 10 pt, conference]{ieeeconf}  

\usepackage{graphicx}
\usepackage{amsmath}
\usepackage{graphicx}
\usepackage{float}
\usepackage{pifont}
\usepackage{pgf}
\usepackage{tikz}
\usepackage{commath}
\usepackage{subfig}
\usepackage{epstopdf}
\usepackage[]{algorithm2e}
\usepackage{algorithmic}
\usepackage{ifthen}
\usepackage{subfig}
\usepackage{bbm}
\usepackage{dsfont}
\usepackage{amssymb}
\usepackage{txfonts}
\usepackage{mathbbol}
\usepackage{multirow}
\usepackage{lipsum}
\usepackage{mathtools}

\IEEEoverridecommandlockouts                              

\title{Flight Control of Sliding Arm Quadcopter with \\ Dynamic Structural Parameters}

\author{Rumit Kumar$^{1}$, Aditya M. Deshpande$^{1}$, James Z. Wells$^{1}$, and Manish Kumar$^{2}$ 
\thanks{$^{1}$R. Kumar, A. M. Deshpande, J. Z. Wells are graduate students at Cooperative Distributed Systems Lab at the University of Cincinnati, Ohio 45221, USA, {\tt\small Email:(kumarrt, deshpaad, wells2jz) @mail.uc.edu}}
\thanks{$^{2}$ M. Kumar is Professor at the Department of Mechanical and Materials Engineering at the University of Cincinnati, Ohio 45221, USA, {\tt\small Email:manish.kumar@uc.edu}} 
\thanks{*Manuscript under review}}

\begin{document}
\maketitle
\pagenumbering{roman}  
\setcounter{page}{1}
\begin{abstract}
The conceptual design and flight controller of a novel kind of quadcopter are presented. This design is capable of morphing the shape of the UAV during flight to achieve position and attitude control. We consider a dynamic center of gravity (\textit{CoG}) which causes continuous variation in a moment of inertia (\textit{MoI}) parameters of the UAV in this design. These dynamic structural parameters play a vital role in the stability and control of the system. 
The length of quadcopter arms is a variable parameter, and it is actuated using attitude feedback-based control law. The \textit{MoI} parameters are computed in real-time and incorporated in the equations of motion of the system. The UAV utilizes the angular motion of propellers and variable quadcopter arm lengths for position and navigation control. The movement space of the \textit{CoG} is a design parameter and it is bounded by actuator limitations and stability requirements of the system. A detailed information on equations of motion, flight controller design and possible applications of this system are provided. Further, the proposed shape-changing UAV system is evaluated by comparative numerical simulations for way point navigation mission and complex trajectory tracking.
\end{abstract}

\section{Introduction}
Multirotor applications in the civilian domain have gained extensive popularity over the past decade. Several companies and researchers are coming up with new designs based on evolving operational requirements. This rapidly growing market demands for innovative design advances to make UAV operations more reliable and safer during the fight. The quadcopters are among the most popular platforms in this space and they come with different weight range based on the payload and endurance requirements during flight. The evolving drone operational requirements pose immense challenges for aircraft designers and flight control engineers.
Tethered UAV configurations, variable blade pitch quadcopters, engine-powered UAVs, tail-sitters, morphological aerial platforms, and tilt-rotor quadcopters are popular names among unique aircraft designs. Here, we discuss a brief overview of these aircraft designs to highlight the diversity and implications of various aerial platforms.

In tethered quadcopters, long-endurance flights can be achieved by using a taut cable to deploy the UAV from ground \cite{nicotra2017nonlinear}. This cable is responsible for continuous power supply to the UAV during flight. The variable blade pitch quadcopter is another interesting design as it can achieve aggressive and inverted flight modes \cite{gupta2016flight}. Unlike conventional multirotor, all propellers in a variable blade pitch UAV spin at the same angular speed whereas navigation is achieved by varying the rotor blade pitch based on the control law. The variable pitch quadcopter utilizes the main motor for driving the propellers and four additional servo motors for controlling the blade pitch during flight. Sheng et al. \cite{sheng2016control} discussed the control and optimization for variable pitch quadcopter and Pang et al. in \cite{pang2016towards} showed a design of gasoline-engine powered variable-pitch quadcopter. The concept of variable pitch quad tilt-rotor (VPQTR) aircraft is discussed in \cite{ferrarese2013modeling}. Tail-sitter UAVs are also popular as they can take off and land like multi-rotors and execute missions like fixed-wing aircraft. The notable work on tail-sitter UAVs was presented in \cite{swarnkar2018biplane} and \cite{ritz2018global}.

Falanga et al. in \cite{8567932} presented morphing quadcopter design which can change shape, squeeze and fly through narrow spaces. This quadcopter can transition between \textit{X, T, H, O} morphology configurations. The shape-changing feature leads to adaptive morphology in UAVs. This extends the flight operations envelope of unmanned aerial systems in an uncertain environment. Aerial manipulation using a shape-changing aerial vehicle is discussed in \cite{zhao2017deformable} and \cite{zhao2018design}. The \textit{DRAGON} can achieve shape transformation in-flight using multiple links and highlight the feasibility of aerial manipulation \cite{zhao2018design}. 
Bucki et al. presented a novel passively morphing design and control of quadrotor \cite{bucki2019design}. This design focused on maneuver such as traversal of the vehicle through small gaps. This design used sprung hinges on the UAV arms which had downward foldable degree of freedom when low thrust commands were applied.
The tilt-rotor quadcopters are over actuated systems with eight servo inputs and provide independent control over each degree of freedom in the system \cite{sridhar2019tilt}. So far, the additional servo inputs in the unmanned aerial systems have been exploited for developing unique morphological configurations, and aerial manipulation. However, we propose a new design aspect by providing active control of the structural parameters of the UAV. The primary objective is to achieve position and attitude control of the UAV by varying the quadcopter arm lengths alone or in combination with rotor speeds.

The structural parameter variation can induce body torques which can be exploited to enable UAV navigation by developing appropriate attitude feedback control laws. Change in UAV shape results in variations of Center of Gravity (\textit{CoG}) and Moment of Inertia (\textit{MoI}) of the system. The structural parameters are computed in real-time to develop an accurate dynamic model. These design changes in vehicle will be useful in augmenting the operational capabilities of the UAVs for different applications such as aerial transportation and heavy payload delivery. There is always a design trade-off between the stability and maneuverability in aerial vehicles. The stability is an important aspect for a large size multirotor UAV. It is difficult to achieve the required control bandwidth in large drones with big propellers by using the conventional motor mixer technique \cite{mulgaonkar2014power}. So, the method of control via-structural parameter variation will prove useful for large size aerial platforms. Similarly, fault-tolerant control in case of a propeller failure is another aspect where this design will prove useful. It has been reported that the tilt-rotor quadcopter can achieve fault-tolerant capabilities in case of a propeller failure during flight \cite{nemati2016stabilizing, kumar2018reconfigurable}. They can complete the flight mission with three functional propellers. Fault tolerance was achieved by the flight controller and the structural reconfiguration in this design. The flight controller reconfiguration logic was implemented at a software level and a passive structural reconfiguration was achieved by changing the quadcopter arm length \cite{sridhar2018fault}. However, these previous works do not account for continuous control of structural reconfiguration parameters. Although work in \cite{wallace2016dynamics} presented a quadrotor design with variable arm length configuration, the control strategy was constrained for symmetric morphing of the drone.
In this paper, we address this limitation by the shape-changing quadcopter design as shown in figure-\ref{fig1}. The continuous control of structural parameters is an important additional capability for control large size drones and for achieving fault-tolerant control in case of propeller failure.

\section{Design and Dynamic Model} \label{sec2}
\begin{figure}[]
	\centering
	\includegraphics[scale=0.35]{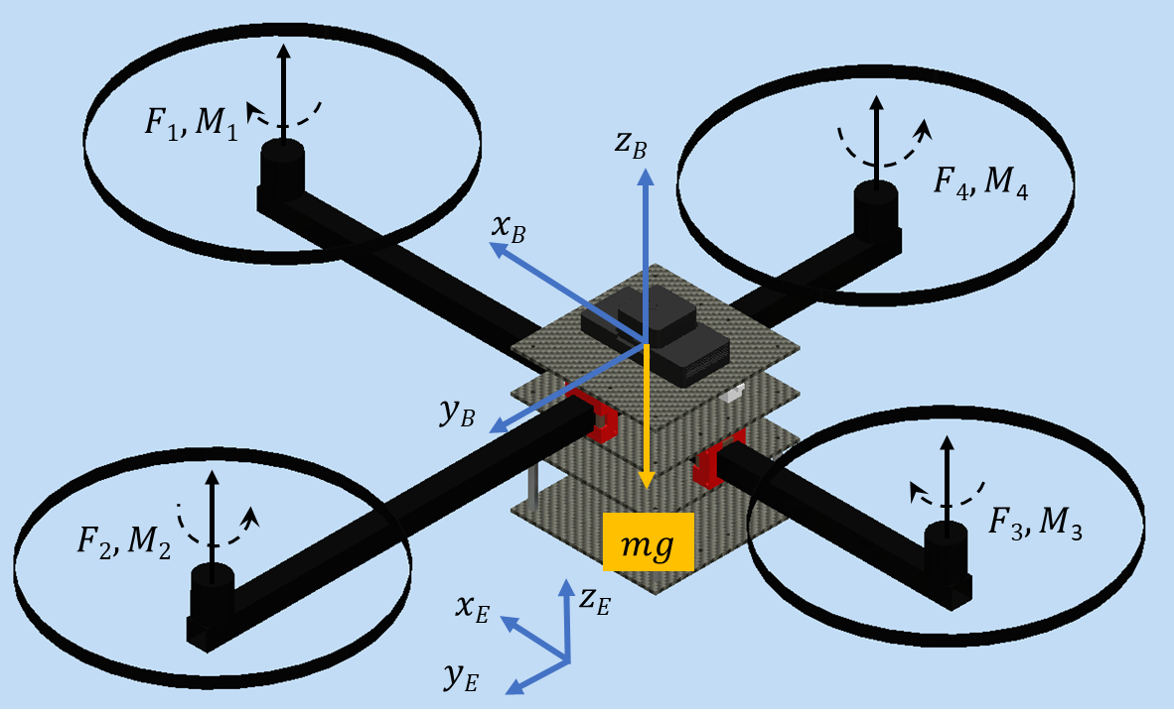}
    \caption{Sliding Arm Quadcopter UAV Design}
    \label{fig1}
\end{figure}

\begin{figure}[h]
	\centering
	\includegraphics[scale=0.3]{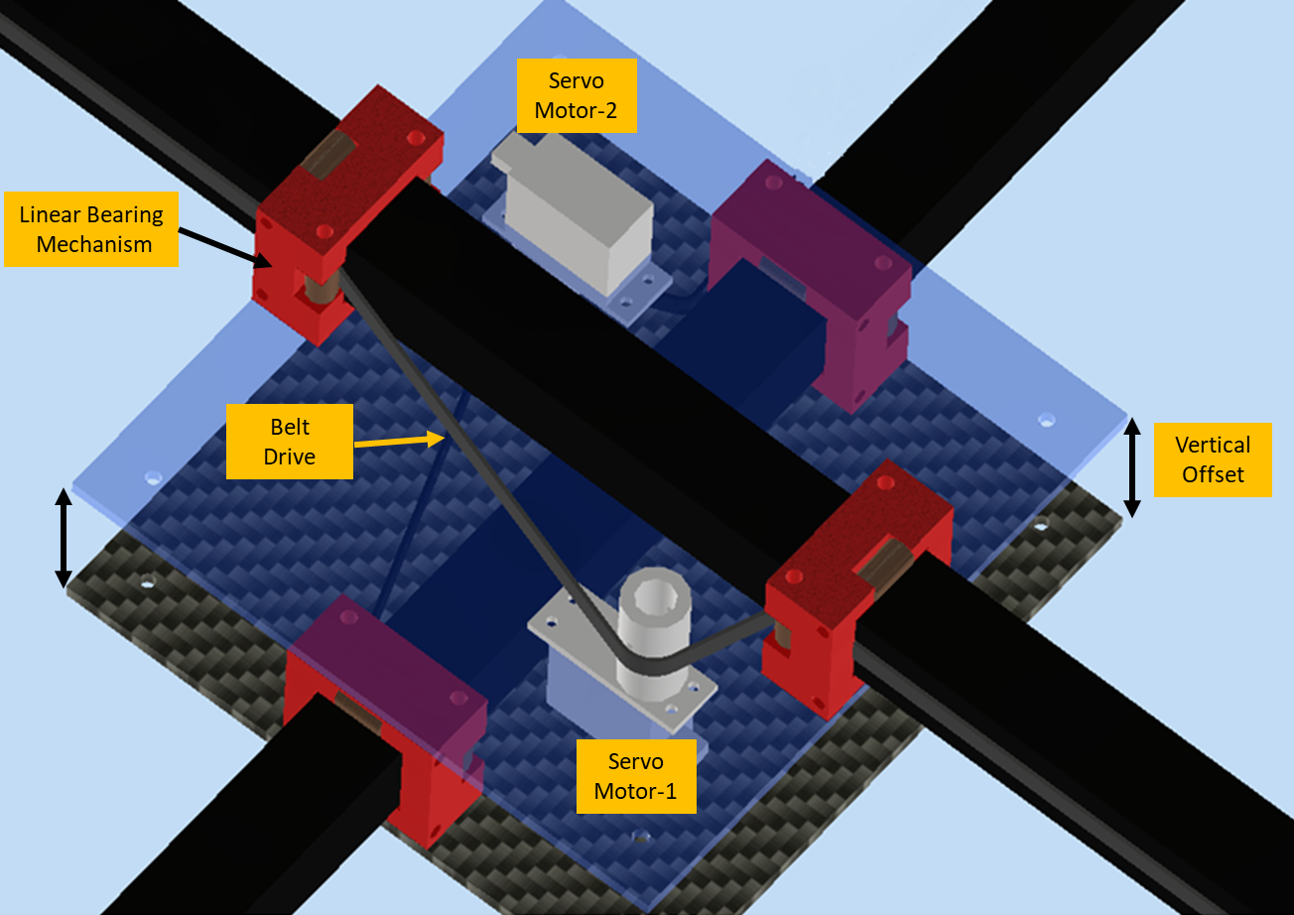}
    \caption{Sliding Arm Mechanism} \vspace{-2mm}
    \label{fig2}
\end{figure}

In this section, the design and mathematical dynamic model of the proposed quadcopter are presented. We consider a \textit{plus} (+) configuration quadcopter as shown in figure \ref{fig1} and the sliding arm mechanism is shown in figure \ref{fig2}. In this design, the system has two long arms with a linear degree of freedom and each arm can slide through the guide bearings. Each arm is independently actuated using a belt-driven servo motor as shown in figure \ref{fig2}. The main motor and propeller combination is present at the ends of each arm such that motor $1$ and $3$ are on one arm and motor $2$ and $4$ are on the other arm to yield a quadcopter configuration. The total length of each arm is constant such that if one rotor moves towards the quadcopter center, the opposite rotor would be moving away and vice-versa. The linear displacement of quadcopter arm with motors $1$ and $3$ is denoted by $\Delta X_l$. Similarly, the linear displacement of quadcopter arm with motors $2$ and $4$ is denoted by $\Delta Y_l$. The system has four motor-propeller pairs and two additional servo motors for actuating the arms. The two arms of this quadcopter have vertical offset which enables their collision-free movement in different planes during the UAV operation.

As discussed earlier, the shape of the sliding arm UAV changes based on the attitude feedback control law. Hence, there is a requirement of a parameter estimation module to compute the $MoI$ matrix in real-time for accurate dynamic modeling and controller development for the system \cite{8567932}. The motion of the UAV is referred w. r. t. the world-frame ($E$) and the body-fixed frame ($B$) moves with the quadcopter. The nominal configuration is defined when the length of all quadcopter arms are equal. In nominal configuration, the system is identical to a hovering conventional quadcopter, the origin of the body frame ($B$) coincides with the $\lbrace CoG \rbrace$ frame and the geometric center of the UAV. In the sliding arm design the $\lbrace CoG \rbrace$ frame and the geometric center of the system can move to a different position in horizontal plane due to structural asymmetry. But, the movement of the $\lbrace CoG \rbrace$ frame does not effect the translational dynamics of the system as shown in \eqref{eq1} and \eqref{eq2}.
\begin{eqnarray}
\begin{bmatrix}
\dot{x}\\
\dot{y}\\
\dot{z}
\end{bmatrix} 
 \hspace{-3mm} &=&  \begin{bmatrix}
c \psi c \theta & c \psi s \theta s \phi-s \psi c \phi & c \psi s \theta c \phi+s \psi s \phi \\
s \psi c \theta & s \psi s \theta s \phi+c \psi c \phi & s \psi s \theta c \phi - c \psi s \phi  \\
-s \theta & c \theta s\phi & c \theta c \phi 
\end{bmatrix} \begin{bmatrix}
u\\
v\\
w
\end{bmatrix} \label{eq1} \\
\begin{bmatrix}
\dot{u}\\
\dot{v}\\
\dot{w}
\end{bmatrix} 
 \hspace{-3mm} &=&  \frac{1}{m}
\begin{bmatrix}
0 \\
0\\
\sum\limits_{i=1}^{4} F_{i}
\end{bmatrix}
- \begin{bmatrix}
-g s\theta\\
g c\theta s\phi\\
g c\theta c\phi
\end{bmatrix}
+
\begin{bmatrix}
rv - qw\\
pw - ru\\
qu - pv
\end{bmatrix}  \label{eq2}
\end{eqnarray}
Here, $m$ and $g$ denote mass of the UAV and acceleration due to gravity. The $sine$ and $cosine$ angle terms are represented as $s(.)$ and $c(.)$ respectively. $[u, v, w]^T$  is the velocity vector in $B$ and $Z-Y-X$ Euler angle transformation is utilized to compute vehicle states in $E$. $[x, y, z]^T$ is the position vector of $B$ in $E$. The total thrust force generated by the propellers is given by $\Sigma F_i, \forall {i} \in \lbrace 1,2,3,4 \rbrace $. The body rate vector is represented by $[p, q, r]^{T}$ in the $\lbrace CoG \rbrace$ frame. 

\begin{figure}[t]
	\centering
	\includegraphics[scale=0.6]{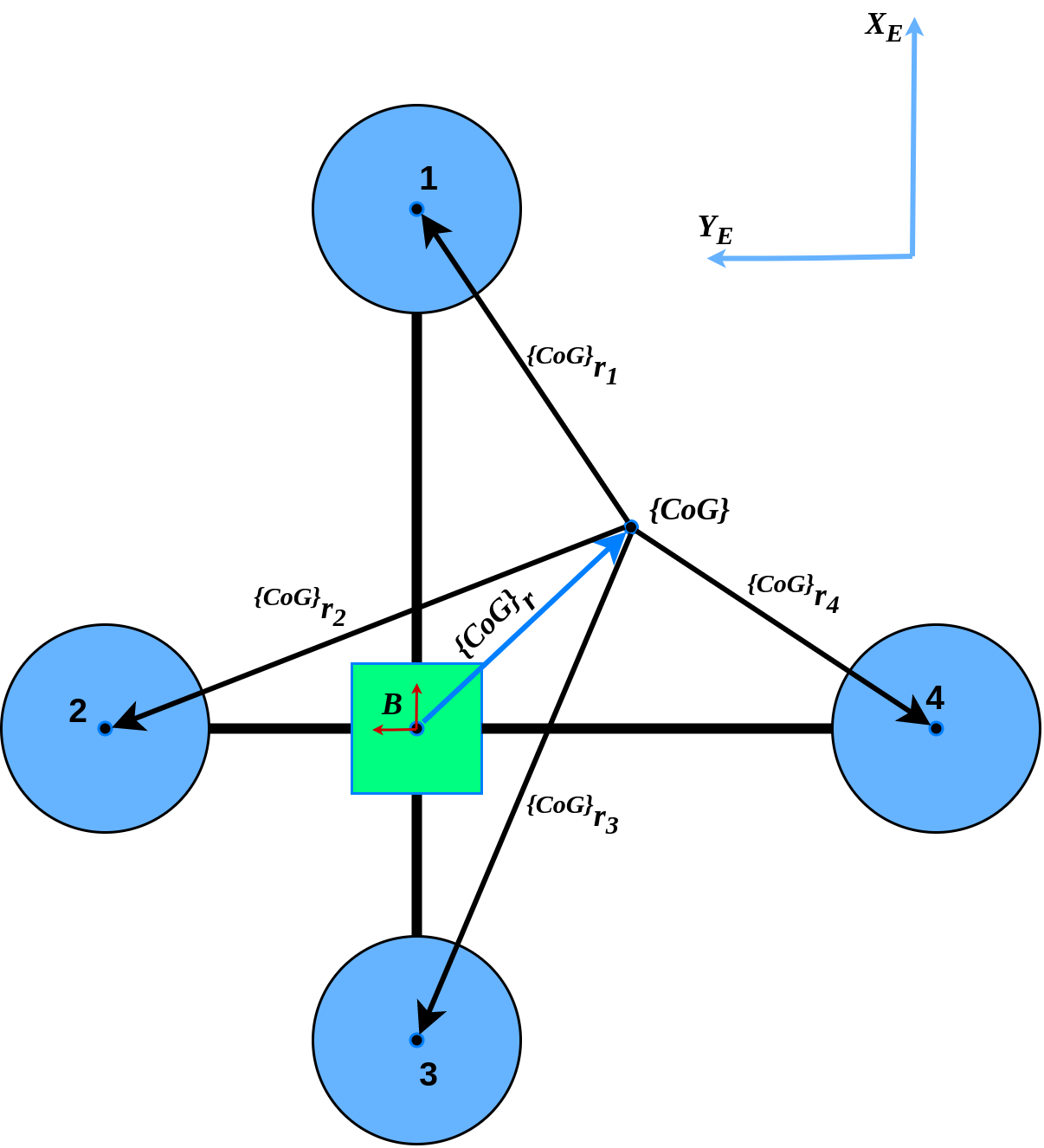}
    \caption{Axes notation for rotational dynamics. \\
    Note: $^{\{CoG\}}r_{i} \equiv (^{\{CoG\}}r_{x_{i}}, ^{\{CoG\}}r_{y_{i}}), \forall {i} \in \{1,2,3,4\}$ \vspace{-6mm} \label{fig3}}
\end{figure}

The rotational dynamics are described similar to \cite{zhao2018transformable} by solving for torque in $\lbrace CoG \rbrace$ frame. A parameter estimation module is used to estimate the instantaneous $MoI$ matrix ($^{\lbrace CoG \rbrace} J$) for the UAV. Firstly, $^{\lbrace CoG \rbrace} r$ vector is computed in $B$. Further, the $MoI$ of each UAV component is transformed to $\lbrace CoG \rbrace$ frame using parallel-axes theorem. The motors and propellers are modelled as cylinders. Similarly, the quadcopter arms and the central body of the UAV are modelled as cuboids. A detailed discussion on skew-symmetric $MoI$ matrix ($^{\lbrace CoG \rbrace} J$) computation is presented in reference \cite{8567932}. It includes computation of $^{\lbrace CoG \rbrace} r$ vector of the system in $B$, and $MoI$ of each component of the UAV such as $^{\lbrace body \rbrace} J$, $^{\lbrace arm \rbrace} J$, $^{\lbrace motor \rbrace} J$, and $^{\lbrace rotor \rbrace} J$. Further, the  $^{\lbrace CoG \rbrace} J$ matrix is obtained by transforming the $MoI$ of the components in ${\lbrace CoG \rbrace}$ \cite{8567932}. The rotational dynamic equations are given by \eqref{eq3}. 
\begin{equation}
^{\lbrace CoG \rbrace} J
\begin{bmatrix} 
\dot{p}\\
\dot{q}\\
\dot{r} 
\end{bmatrix} = \begin{bmatrix} 
\sum\limits_{i=1}^{4}{^{\lbrace CoG \rbrace} r_{y_i} F_{i}} \\
- \sum\limits_{i=1}^{4}{^{\lbrace CoG \rbrace} r_{x_i} F_{i}} \\
 \sum\limits_{i=1}^{4}{(-1)^{i} M_{i}}
\end{bmatrix} 
-
\begin{bmatrix} 
p\\
q\\
r
\end{bmatrix}
\times ^{\lbrace CoG \rbrace} J
\begin{bmatrix} 
p\\
q\\
r
\end{bmatrix} \label{eq3}
\end{equation}
Here, $F_i$, and $M_i$, $\forall {i}\in \lbrace1,2,3,4\rbrace$ are the force and moment produced by the respective propeller and they are directly proportional to the squared angular speed ($\omega$) of the propeller as described in \cite{grasp}. The axes notation for the rotational dynamics are highlighted in figure \ref{fig3}. It should be noted that the equation \eqref{eq3} has components of torque from all motors for each rotational degree of freedom in the UAV. The torque produced by any motor about $\lbrace CoG \rbrace$ frame is dependent on  $^{\lbrace CoG \rbrace} r$ vector and angular speed ($\omega$) of the motor. The Euler angles for orientation representation can be computed by integrating the standard Euler angle rate equation \cite{grasp}.
It is considered that the body frame $B$ and $\lbrace CoG \rbrace$ frame will have the same orientation during flight \cite{zhao2018transformable}. Thus, the body rates $[p, q, r]^T$ can be directly measured using an inertial measurement unit (IMU) sensor which can be utilized to estimate the orientation of the system.

\section{Controller Development}
In this section, the controller development strategy for the sliding arm quadcopter is presented. This system has six motor inputs. The dynamic model presented in section-\ref{sec2} is used for controller design. The control system of the sliding arm UAV is based on the conventional cascaded loop architecture. It consists of an outer position control module which generates orientation commands for the inner attitude controller. An additional parameter estimation module is implemented for updating the $MoI$-matrix to account for change in the shape of UAV.
\subsection{Position Control}
The position control is the outer loop of the controller and it generates attitude set point commands for the inner attitude controller of the UAV. The position state feedback in world frame is utilized to compute position and velocity errors ($e_x, e_y, e_z, \dot{e_x}, \dot{e_y}, \dot{e_z}$). This information is utilized  by the PID controller loop to compute desired accelerations commands $\ddot{r_i}^{des}; \forall {i}\in \lbrace{x,y,z}\rbrace$ for the UAV as shown in \eqref{eq5}. The acceleration due to gravity is compensated as a feed-forward term in the $z_E$-controller.
\begin{eqnarray}
\nonumber \ddot{r_x}^{d} &=& k_{p_x}e_x + k_{i_x}\int{e_x} dt +  k_{d_x}\dot{e_x} \\
\ddot{r_y}^{d} &=& k_{p_y}e_y + k_{i_y}\int{e_y} dt +  k_{d_y}\dot{e_y} \label{eq5} \\ 
\nonumber \ddot{r_z}^{d} &=& k_{p_z}e_z + k_{i_z}\int{e_z} dt +  k_{d_z}\dot{e_z}  + g
\end{eqnarray} 
Here, $k_{p_i}$, $k_{i_i}$, and $k_{d_i}$ $\forall {i}\in \lbrace{x,y,z}\rbrace$ are the proportional, integral and derivative gains for the position controller. The rotor angular speed required for individual propeller motors necessary for hovering and motion along the $z_E-$axis is given by $\omega_h$ as shown in \eqref{eq6}.
\begin{eqnarray} 
\omega_h=\sqrt{\dfrac{m\ddot{r_z}^{d}}{4k_f}}   \label{eq6}
\end{eqnarray}
The desired accelerations along $x_E y_E$-axes from (\ref{eq5}) are used to compute the attitude set point commands. The pitch and roll angle set point commands for the UAV are represented as  ${\theta}^{d}$ and ${\phi}^{d}$ respectively as shown in \cite{grasp}.
\begin{eqnarray}
{\phi}^{d} &=& \frac{\ddot{r_x}^{d}sin{\psi}^{d}-\ddot{r_y}^{d}cos{\psi}^{d}}{g}
      \label{eq7}
\\
\nonumber\\
{\theta}^{d} &=& \frac{\ddot{r_x}^{d}cos{\psi}^{d}-\ddot{r_y}^{d}sin{\psi}^{d}}{g}
      \label{eq8}
\end{eqnarray}
where, $\psi^{d}$ is the desired yaw angle for the system. The attitude set point commands are sent to the attitude controller. 
\subsection{Attitude Control}
In conventional quadcopter, the attitude control loop generates rotor angular speed commands $\omega_i, \forall {i}\in \lbrace{1,2,3,4}\rbrace$ in the form of PWM signals to the UAV motors. However, the sliding arm quadcopter design has two additional servo inputs for actuating the quadcopter arms. The linear displacement of quadcopter arms are represented by $\Delta X_l$ and $\Delta Y_l$. The attitude set point commands are given by \eqref{eq7} and \eqref{eq8} and the attitude state feedback is utilized to compute the error ($e_\phi, e_\theta, e_\psi, e_p, e_q, e_r$) in vehicle orientation with respect to the attitude commands. This information is utilized  by the PID controller to compute the change in the angular speed ($\Delta \omega_i,  \forall {i}\in \lbrace{\phi, \theta, \psi}\rbrace$) of the rotors and quadcopter arm displacement ($\Delta X_l, \Delta Y_l$) as shown in \eqref{eq9}.
\begin{eqnarray}
\nonumber \Delta \omega_i &=& k_{p_i}e_i + k_{i_i}\int{e_i} dt +  k_{d_i}\dot{e_i} ; \quad \forall {i}\in \{\phi, \theta, \psi\} \\
\Delta X_l &=& k_{p_{l_{\theta}}}e_\theta + k_{i_{l_{\theta}}}\int{e_\theta} dt +  k_{d_{l_{\theta}}}\dot{e_\theta}\\ \label{eq9}
\nonumber \Delta Y_l &=& k_{p_{l_{\phi}}} e_\phi + k_{i_{l_{\phi}}} \int{e_{\phi}} dt +  k_{d_{l_{\phi}}} \dot{e_{\phi}} 
\end{eqnarray} 
Here, $k_{p_i}$, $k_{i_i}$, and $k_{d_i}$ $\forall {i}\in \lbrace{\phi, \theta, \psi, l_\theta, l_\phi}\rbrace$ are the proportional, integral and derivative gains for the attitude controller. The angular speeds outputs ($\Delta \omega_i,  \forall {i}\in \lbrace{\phi, \theta, \psi}\rbrace$) are passed to the motor mixing module to generate rotor commands as shown in \cite{grasp}. Similarly, the linear displacement commands ($\Delta X_l, \Delta Y_l$) are used in servo motors governing the sliding motion of the quadcopter arms. 
\begin{figure}[b]
	\centering
	\includegraphics[scale=0.6]{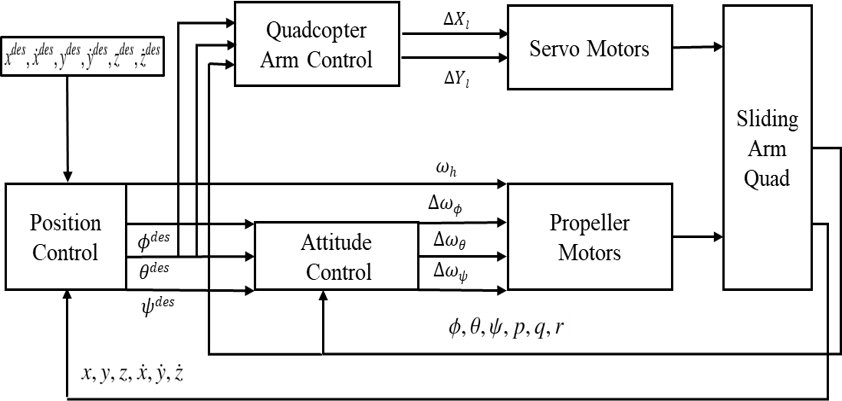}
    \caption{Control Architecture}
    \label{fig4}
\end{figure}
The complete signal flow and control architecture of the sliding arm UAV is shown in figure \ref{fig4}. The servo motors for actuating the quadcopter arms are modelled by classic second order transfer function as shown in equation \eqref{eq11}.
\begin{equation}
G(s) = \frac{\omega_n^2}{s^{2}+2\zeta\omega_n + \omega_n^2}
      \label{eq11}
\end{equation}
In the actual system, a servo motor with higher bandwidth would be necessary for faster actuation of quadcopter arms. There are commercially available digital servo motors with an operational frequency of up to $15 rad/s$ which would be an ideal choice for this application \cite{servo_db}.

\section{Numerical Simulations and Results}
In this section, the proposed controller is validated by numerical simulations. The mathematical model of the UAV and controller are developed in MATLAB and Simulink R2017a. Two types of numerical simulations were considered in this work. The parameters used in the simulations are total mass of the system $m=1.56kg$, nominal quadcopter arm length $l=0.25m$, thrust and moment coefficients for the propellers are $k_f=2.2e-4 Ns/rad$, and $k_m=5.4e-6Ns/rad$ respectively and the $MoI$-matrix is a variable quantity. The first simulation shows the performance of the system in a conventional way point navigation mission. The second simulation shows the performance of the UAV for complex trajectory tracking in the presence of sensory noise. 

\subsection{Way Point Navigation Simulation}
\begin{figure}[]
	\centering
	\includegraphics[scale=0.4]{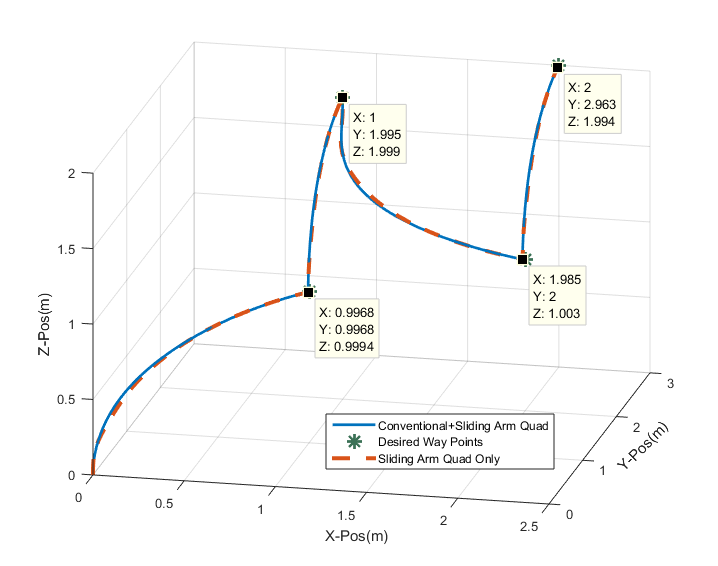}
    \caption{Three dimensional Trajectory} \vspace{-2mm}
    \label{fig5}
\end{figure}

\begin{figure}[]
	\centering
	\includegraphics[scale=0.4]{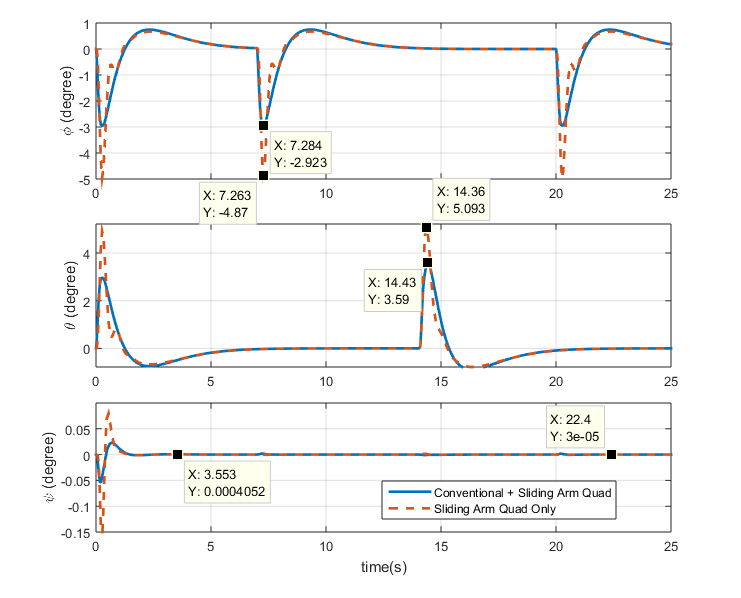}
    \caption{Variation in Euler angles} \vspace{-4mm}
    \label{fig6}
\end{figure}

\begin{figure}[]
	\centering
	\includegraphics[scale=0.4]{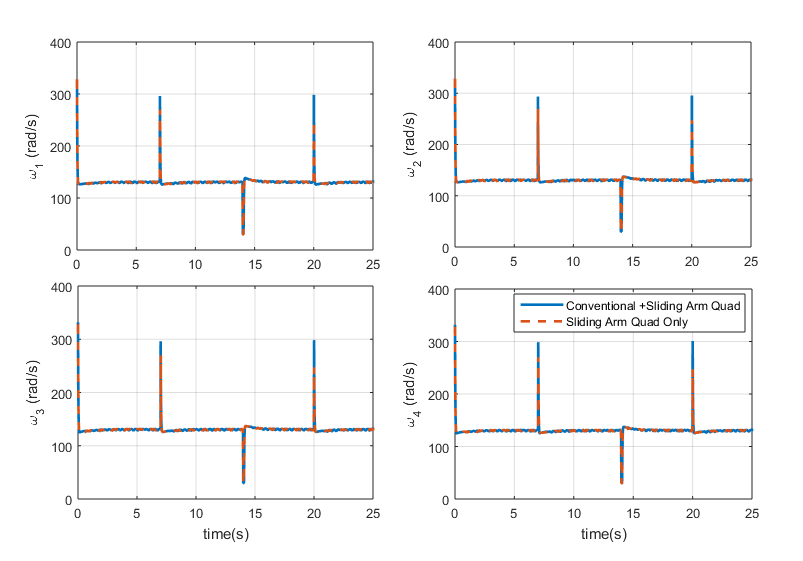}
    \caption{Variation in angular speed of rotors} \vspace{-4mm}
    \label{fig7}
\end{figure}

\begin{figure}[]
	\centering
	\includegraphics[scale=0.4]{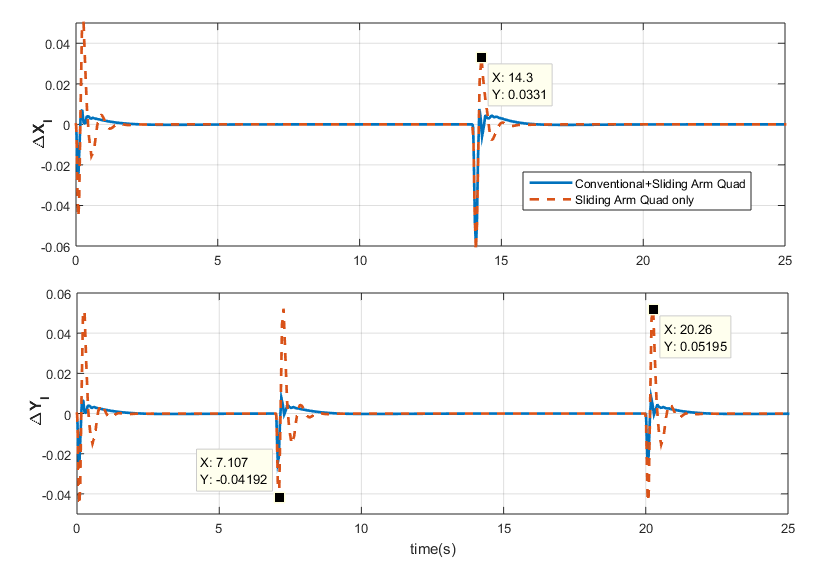}
    \caption{Variation in quadcopter arm lengths ($\Delta X_l, \Delta Y_l$)} \vspace{-4mm}
    \label{fig8}
\end{figure}

The UAV is initialized at the origin and commanded to visit a predefined set of way points given by $\{(x_d, y_d, z_d)\}: \{(1,1,1), (1,2,2), (2,2,1), (2,3,2)\}$. The dimensions are in meters. The objective of this study is to assess the performance of the control scheme and novel quadcopter design for two cases: (i) the attitude controller is governed by the combination of variation in propeller rpm ($\Delta \omega_i,  \forall {i}\in \lbrace{\phi, \theta, \psi}\rbrace$) and quadcopter sliding arm control ($\Delta X_l, \Delta Y_l$) simultaneously (called `Conventional + Sliding Arm Quad' in the figures); (ii) the attitude controller is solely governed by the quadcopter sliding arm ($\Delta X_l, \Delta Y_l$) and propeller rpm are not varied except for thrust and yaw control such that ($\Delta \omega_\phi = \Delta \omega_\theta = 0$) (called `Sliding Arm Quad Only' in the figures). The three-dimensional trajectory of the UAV is shown in figure \ref{fig5}. The UAV can visit the set of way points successfully in both cases. Figure \ref{fig6} shows the variation of Euler angles during flight. It can be seen that sliding arm quadcopter has higher transient peaks while changing direction across way point. However, the combination based controller achieves smoother transition across way points. Figure \ref{fig7} shows the variation in the angular speed of the propellers during flight. Figure \ref{fig8} shows the variation in quadcopter sliding arm length for both the cases. As observed earlier, the sliding arm control has larger transient peaks while functioning alone for controlling the attitude of the UAV. However, the transient peaks are minimized in the combination based controller. The overall performance of both systems is very similar, but it is interesting to note that the proposed system can navigate just by changing the shape of the UAV during flight. The flying characteristics are also very comparable to the conventional quadcopter. This can be attributed to the dynamic nature of the center of gravity resulting in control torques for navigating the UAV. The sliding arm quadcopter design is useful in the absence of the conventional rotor angular speed controller such as for the large size multirotors.  
\subsection{Trajectory Tracking Simulation}
\begin{figure}[]
	\centering
	\includegraphics[scale=0.35]{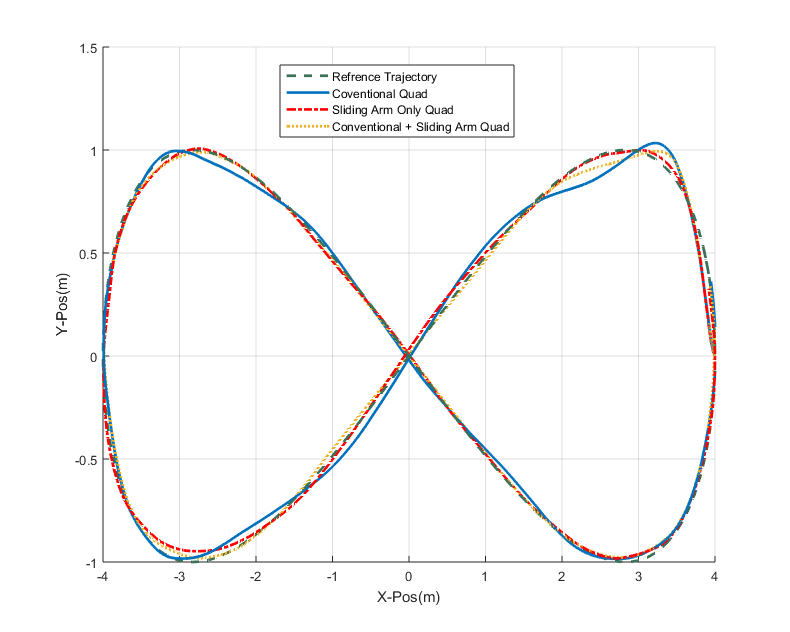}
    \caption{Two dimensional Trajectory} \vspace{-4mm}
    \label{fig10}
\end{figure}

\begin{figure}[]
	\centering
	\includegraphics[scale=0.4]{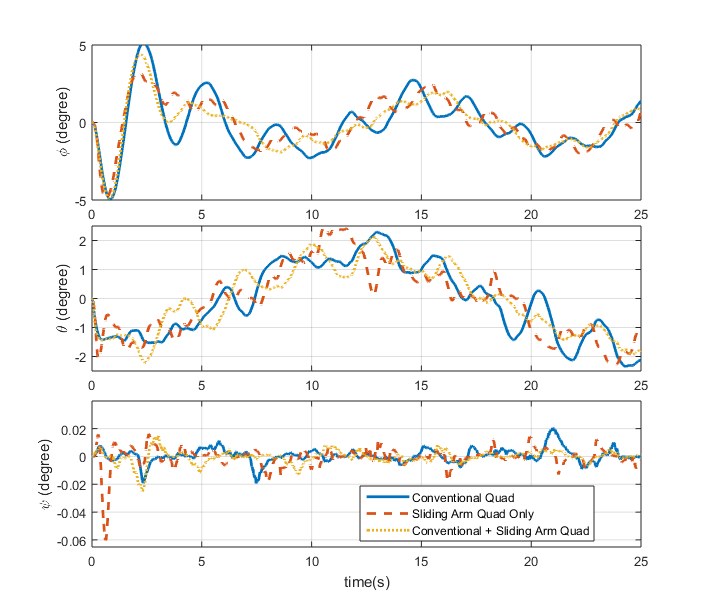}
    \caption{Variation in Euler angles} \vspace{-4mm}
    \label{fig11}
\end{figure}

\begin{figure}[]
	\centering
	\includegraphics[scale=0.4]{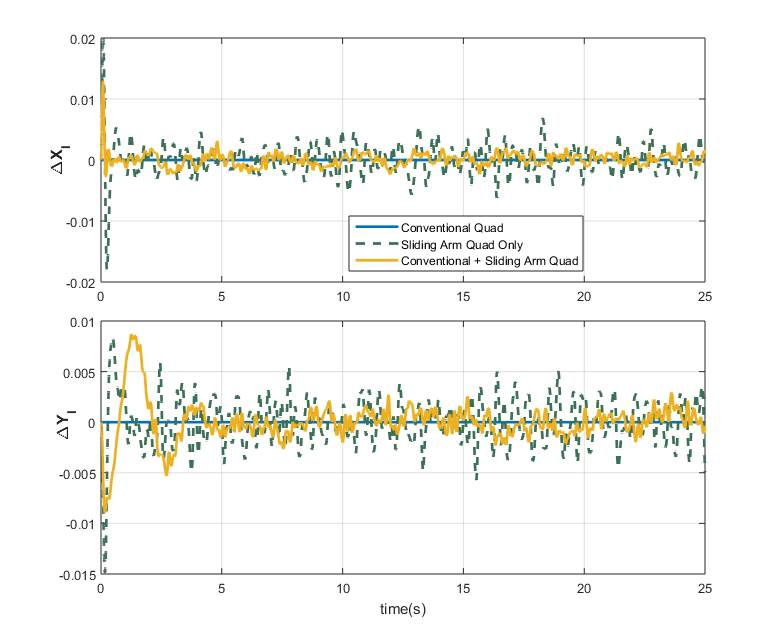}
    \caption{Variation in quadcopter arm lengths ($\Delta X_l, \Delta Y_l$)} \vspace{-4mm}
    \label{fig12}
\end{figure}

Here, the performance of the system is evaluated similar to the previous case. The UAV is commanded to track a figure-eight trajectory. The equations governing the desired flight path are the same as described in \cite{kumar2017tilting}. The simulation is performed with uncertainties in the roll and pitch angles of the UAV. A uniformly distributed random noise is considered in the range of $-2$ to $+2$ degrees. The simulation is performed for three cases: (i) a conventional quadcopter configuration with no sliding arm functionality (called `Conventional Quad' in the figures); (ii) the attitude controller is solely governed by the quadcopter sliding arm ($\Delta X_l, \Delta Y_l$) and ($\Delta \omega_\phi = \Delta \omega_\theta = 0$); (iii) the attitude controller is governed by the combination of variation in propeller rpm and quadcopter sliding arm control simultaneously. Figure \ref{fig10} shows the tracking characteristics of these three cases. It can be seen that conventional quadcopter experiences a larger transient response. There is also a lateral movement in the tracking characteristics of the conventional quadcopter due to uncertainties. However, the system with a sliding arm mechanism has a lower and well-damped transient response. It tracks the desired trajectory very closely with only a sliding arm mechanism as well as in a combination based controller. The sliding arm mechanism assists to increase the trajectory tracking performance. Figure \ref{fig11} shows the variation of Euler angles in the three cases during flight. The magnitude of the roll angle is more in conventional quadcopter as compared to the sliding arm system. This justifies the lateral movement in the conventional quadcopter. The pitch and yaw angles are very small in both systems. Figure \ref{fig12} shows the active control and displacement of quadcopter arm during flight. As the UAV tracks the desired trajectory the active control of the sliding arm mechanism is instrumental in providing disturbance rejection. 
 \vspace{-3mm}
\section{Conclusion}
In this paper, the conceptual design and flight controller for the sliding arm UAV were presented. A preliminary design of the shape-changing UAV by sliding arm mechanism was described. The equations of motion for translational and rotational dynamics were discussed by considering a dynamic center of gravity. This required online parameter estimation for developing an accurate dynamic model. The flight controller architecture was presented for the sliding arm UAV. The proposed controller was evaluated against conventional quadcopter by numerical simulation for way point navigation and complex trajectory tracking. It was observed that the proposed system navigated through all way points by changing the shape. The dynamic nature of the center of gravity resulted in generating control torques for navigating the UAV. The active control of the sliding arm mechanism is shown to exhibit enhanced disturbance rejection capability and maneuverability. This system would be useful for controlling large size multirotor platforms and achieving fault-tolerant control. Future work will involve the development of the experimental prototype and flight testing of the proposed sliding arm UAV.
\bibliographystyle{IEEEtran}
\bibliography{main}
\end{document}